\documentclass[acmsmall]{acmart}
\usepackage{booktabs} % For formal tables

\usepackage[ruled]{algorithm2e} % For algorithms

\usepackage{caption}
\usepackage{subcaption}
\usepackage{graphicx}
\usepackage{multirow}
\usepackage{float}
\usepackage{subcaption}
\usepackage{lineno,hyperref}
\modulolinenumbers[5]
\usepackage{algorithmic}
\usepackage{multirow}
\usepackage{amsmath}
\SetAlFnt{\small}
\SetAlCapFnt{\small}
\SetAlCapNameFnt{\small}
\SetAlCapHSkip{0pt}
\IncMargin{-\parindent}

\copyrightyear{2018}
\acmYear{2018}
\setcopyright{acmlicensed}
\acmConference[Woodstock '18]{Woodstock '18: ACM Symposium on Neural Gaze Detection}{June 03--05, 2018}{Woodstock, NY}
\acmBooktitle{Woodstock '18: ACM Symposium on Neural Gaze Detection, June 03--05, 2018, Woodstock, NY}
\acmPrice{15.00}
\acmDOI{10.1145/1122445.1122456}
\acmISBN{978-1-4503-9999-9/18/06}

\begin{document}

%
% The "title" command has an optional parameter, allowing the author to define a "short title" to be used in page headers.
\title{Explainable Event Recognition}
\author{Imran Khan}
 \email{imran.cse@uetpeshawar.edu.pk}
\affiliation{%
  \institution{University of Engineering and Technology, Peshawar, Pakistan}
}

\author{Kashif Ahmad}
\email{kahmad@hbku.edu.qa}
\affiliation{%
  \institution{College of Science and Engineering (CSE), Hamad Bin Khalifa University, Doha}
}
\author{Namra Gul}
\email{nmgul.msee20seecs@seecs.edu.pk}
\affiliation{%
  \institution{School of Electrical Engineering and Computer Sciences, NUST Islamabad, Pakistan}
 }

\author{Talhat Khan}
\email{talhat.cse@uetpeshawar.edu.pk}
\affiliation{%
 \institution{University of Engineering and Technology, Peshawar, Pakistan}
 }
 \author{Nasir Ahmad}
 \email{n.ahmad@uetpeshawar.edu.pk}
\affiliation{%
  \institution{University of Engineering and Technology, Peshawar, Pakistan}
  }
\author{Ala Al-Fuqaha}
\email{aalfuqaha@hbku.edu.qa}
\affiliation{
  \institution{College of Science and Engineering (CSE), Hamad Bin Khalifa University, Doha, Qatar.}
  }

%
% By default, the full list of authors will be used in the page headers. Often, this list is too long, and will overlap
% other information printed in the page headers. This command allows the author to define a more concise list
% of authors' names for this purpose.
\renewcommand{\shortauthors}{Khan et al.}

%
% The abstract is a short summary of the work to be presented in the article.
\begin{abstract}
The literature shows outstanding capabilities for CNNs in event recognition in images. However, fewer attempts are made to analyze the potential causes behind the decisions of the models and exploring whether the predictions are based on event-salient objects/regions? To explore this important aspect of event recognition, in this work, we propose an explainable event recognition framework relying on Grad-CAM and an Xception architecture-based CNN model. Experiments are conducted on three large-scale datasets covering a diversified set of natural disasters, social, and sports events. Overall, the model showed outstanding generalization capabilities obtaining overall F1-scores of 0.91, 0.94, and 0.97 on natural disasters, social, and sports events, respectively. Moreover, for subjective analysis of activation maps generated through Grad-CAM for the predicted samples of the model, a crowd-sourcing study is conducted to analyze whether the model's predictions are based on event-related objects/regions or not? The results of the study indicate that 78\%, 84\%, and 78\% of the model decisions on natural disasters, sports, and social events datasets, respectively, are based on event-related objects/regions.
\end{abstract}

%\begin{CCSXML}
%<ccs2012>
 %<concept>
  %<concept_id>10010520.10010553.10010562</concept_id>
  %<concept_desc>Computer systems organization~Embedded systems</concept_desc>
  %<concept_significance>500</concept_significance>
 %</concept>
 %<concept>
  %<concept_id>10010520.10010575.10010755</concept_id>
  %<concept_desc>Computer systems organization~Redundancy</concept_desc>
  %<concept_significance>300</concept_significance>
 %</concept>
 %<concept>
  %<concept_id>10010520.10010553.10010554</concept_id>
  %<concept_desc>Computer systems organization~Robotics</concept_desc>
  %<concept_significance>100</concept_significance>
 %</concept>
 %<concept>
  %<concept_id>10003033.10003083.10003095</concept_id>
  %<concept_desc>Networks~Network reliability</concept_desc>
  %<concept_significance>100</concept_significance>
 %</concept>
%</ccs2012>
%\end{CCSXML}

%\ccsdesc[500]{Computer systems organization~Embedded systems}

%
% Keywords. The author(s) should pick words that accurately describe the work being
% presented. Separate the keywords with commas.

%
% A "teaser" image appears between the author and affiliation information and the body 
% of the document, and typically spans the page. 
%%  \includegraphics[width=\textwidth]{sampleteaser}
%%  \caption{Seattle Mariners at Spring Training, 2010.}
%%  \Description{Enjoying the baseball game from the third-base seats. Ichiro Suzuki preparing to bat.}

%
% This command processes the author and affiliation and title information and builds
% the first part of the formatted document.
\maketitle

\section{Introduction}
\label{sec:introduction}
Event recognition in multimedia content (i.e, videos, images, sound tapes, and textual information) has been widely explored in the literature \cite{ahmad2019deep}. Several interesting definitions and aspects of events have emerged as part of the efforts in different application domains. The definition of an event largely depends on the nature of the application and the type of content to be analyzed. For instance, in the complex videos, an event can be defined through interactions of multiple objects or/\& human actions in scenes with various types of backgrounds \cite{gan2015devnet}. In audio tapes, events depict the occurrence of a certain sound pattern \cite{chandrakala2021multi}. In images, on the other hand, events are represented and differentiated through the presence of certain objects and scenes \cite{ahmad2019deep}. 

A vast majority of the literature aims at event recognition in images where mostly the focus remained on the detection and recognition of certain objects and background information. For instance, in \cite{wang2015object}, object and scene-level features extracted through individual Convolutional Neural Networks (CNNs), pre-trained on ImageNet \cite{deng2009imagenet} and Places \cite{zhou2014learning} datasets, are combined for event recognition. However, all the objects and scenes in an image do not necessarily belong to an underlying event. In fact, there are certain objects and scenes that define and differentiate events. These objects and scenes may lie at any location covering a large or a smaller portion of an image, and are not necessarily be the most prominent parts of the image \cite{rosani2015eventmask,ahmad2018saliency}. %The hypothesis is proved true by Rosani et al. \cite{rosani2015eventmask} by introducing the concept of event saliency. %To this aim, a large population is involved in a game-based framework for crowd-sourcing to extract/identify event-related objects and scenes from image.  

Despite such challenges, Deep Learning (DL) models have shown outstanding generalization capabilities in event recognition. The average performance of state-of-the-art methods in terms of average accuracy on widespread datasets, such as \textit{culture}, \textit{social}, \textit{sports}, and \textit{natural disasters} events datasets are in the range of 87\%, 70\%, 85\%, and 68\%, respectively \cite{ahmad2019deep}.  However, the end-to-end learning mechanism of DL models without adequate explanation of the predictions leads to several questions, such as \textit{which parts (i.e., objects and scenes) of an image contribute more to the prediction of the model}? \textit{are these parts/regions of the images contain event-related objects and scenes}? \textit{Do all the decisions of the models are based on event-salient objects/regions?} \textit{Do they make sense to humans}?

In this work, we aim to answer these questions by extending the concept of explanaility to event recognition. To this aim, we propose an explainable event recognition framework relying on Gradient-weighted Class Activation Mapping (Grad-CAM) and a CNN architecture. Moreover, the experiments are conducted on three different datasets covering social, sports, and natural disaster events. We believe such a detailed analysis of a diversified set of events will provide a baseline for future work in the domain.

The main contributions of the work can be summarized as follows:

\begin{enumerate}
    \item We extend the concept of explainability to event recognition by adopting explainability methods for a CNN-based event recognition framework.
    \item We also conduct a crowd-sourcing study to analyze whether the model's predictions are made on event salient objects/regions or not? The study will allow us to analyze how much the models consider event-salient objects/regions in prediction? 
\item Through an extensive experiments on three different datasets, we aim to answer the following research questions:
\begin{enumerate}
    \item Which parts (i.e., objects and scenes) of an image contribute more to the prediction of the model? 
    \item Are these parts of the images contain event-related objects and scenes? 
    \item Are all the predictions of the model based on event salient objects and regions? 
    \item Do they make sense for humans?
\end{enumerate}
\end{enumerate}

The rest of the paper is organized as follows. Section \ref{sec:related_work} provides an overview of literature on event recognition. Section \ref{sec:methodology} describes the proposed framework. Section \ref{sec:dataset} summarizes the characteristics and statistics of the datasets used for the experimentation. Section \ref{sec:experments} reports the experimental results and key lessons learned. Finally, Section \ref{sec:conclusion} provides some concluding remarks. 

\section{Related Work}
\label{sec:related_work}
The literature explores different aspects of event recognition in images. For instance, several types of events, such as social \cite{ahmad2016used}, cultural \cite{baro2015chalearn}, political \cite{ahsan2017complex}, sports \cite{li2007and}, and natural disaster events \cite{said2019natural}, are reported in the literature. From the implementation point of view, different strategies have been adopted relying on both visual and metadata information whenever available. Though the additional information has been proved effective in several studies, the main focus remained on visual content mostly relying on DL algorithms \cite{ahmad2019deep}. Moreover, DL techniques are mostly employed in three different ways either (i) fine-tuning existing pre-trained models, (ii) training or re-training a DL model, and (iii) extracting features with pre-trained models. For instance, Liu et al. \cite{liu2015exploiting} fine-tuned two pre-trained models namely VggNet \cite{simonyan2014very} and GoogLeNet \cite{szegedy2015going} on a cultural events dataset. In \cite{park2015cultural}, a CNN model composed of three training and pooling, and two fully connected layers is trained on image regions extracted from cultural events images. In \cite{ahmad2017pool}, multiple deep models are employed for feature extraction from event-related images. 

In visual content-based approaches, the focus remained on objects and scenes (i.e., backgrounds), as a result, several interesting event recognition frameworks, employing object and scene-level features both individually and jointly, are proposed. For instance, Wang et al. \cite{wang2018transferring} combined object and scene-level features extracted through DL models in three different ways using initialization, knowledge, and data-based transfer learning. Similarly, in \cite{wei2015deep} CNN models pre-trained on objects \cite{deng2009imagenet} and scenes datasets \cite{zhou2014learning} are combined in late and early fusion. However, events are represented and differentiated by certain objects and scenes, which are not necessarily be the most prominent parts of an event-related image. This hypothesis is proved by Rosani et al. \cite{rosani2015eventmask} through a game-based crowdsourcing approach where a large population was asked to highlight/mask the image parts representing an underlying event.   

Despite the challenges in the identification/extraction of the event-related objects in an image, end-to-end DL methods have shown outstanding performance. However, the literature is still missing a detailed analysis of what kind of image parts contribute more to the decisions of these models. In this work, we aim to explore the causes behind the predictions made by these models and analyze whether the decisions make sense to humans or not.  

\section{Methodology}
\label{sec:methodology}
\begin{figure}[h]
%\label{fig:taxonomy}
\centering
\includegraphics[width=0.75\textwidth]{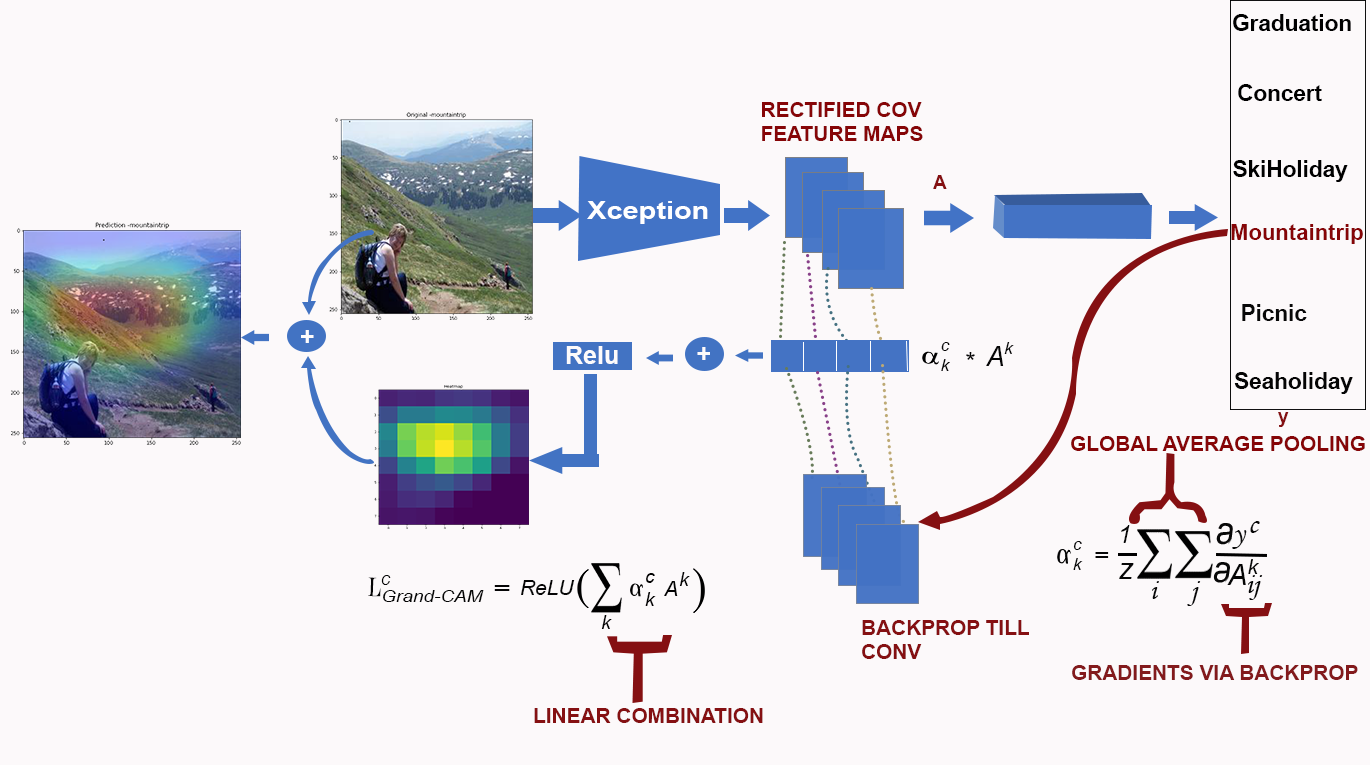}
\caption{Block diagram of the proposed methodology.}
	\label{fig:methodology}
\end{figure}

Figure \ref{fig:methodology} provides the block diagram of the proposed methodology. The proposed framework for explainable event recognition is composed of two main components. The first component is consisted of fine-tuning a pre-trained CNN model while the other component is based on the Grad-CAM, which generates the activation maps of the images processed by the CNN model. Our CNN model is based on a state-of-the-art architecture namely Xception \cite{chollet2017xception}, pre-trained on a large-scale object recognition dataset known as ImageNet \cite{russakovsky2015imagenet}. In the next subsection, we provide details of each component of the framework.

%%%%%%%%%%%%%%%%%%%%%%%%%%%%%%%%%%%%%%%%%
%%%%%%%%%%%%%%%%%%%%%%%%%%%%%%%%%

\subsection{Fine-tuning the Model}
%Our CNN model is based on Xception architecture \cite{chollet2017xception}, which is known as an extreme version of inception and is based on depthwise separable convolution layers. 
The Xception architecture is mainly based on the decoupling hypothesis of mapping cross-channel and spatial correlations. In other words, spatial convolution is performed on each input channel independently. Subsequently, pointwise (i.e., 1x1) convolutional layers are used to project the output of the depthwise convolution. In total, the network consists of 36 convolutional layers, which are structured into 14 modules. All the modules have linear residual connections.

In this work, we fine-tuned a pre-trained version of the model trained on ImageNet dataset \cite{russakovsky2015imagenet}. ImageNet is a large-scale object recognition dataset composed of 1000 classes. Fine-tuning brings several advantages. On one side, it reduces training time and computational resources. On the other hand, improves the performance on smaller dataset lacking a sufficient number of training samples. The parameters used in fine-tuning the model are detailed in Section \ref{sec:experments}. 

\subsection{Grad-CAM Visualization}
In this work, for heatmap visualization, we rely on Gradient-weighted Class Activation Mapping (Grad-CAM) algorithm \cite{selvaraju2017grad}, which is a modified version of CAM \cite{zhou2016learning}. CAM allows identifying the salient (i.e., discriminative) objects/regions of an image provided to a CNN model by computing class activation maps. The algorithm assigns an importance score to every region by computing a weighted combination of the activations. To this aim, all the fully connected layers from the model are replaced with a global average pooling layer, which helps to avoid over-fitting. The average pooling layer is then followed by a classification (i.e., soft-max) layer. 

Grad-CAM extends CAM by generalizing the concept to general CNN architectures by utilizing gradient information in assigning weights to the feature maps. To be more specific the gradient of the loss w.r.t. the last convolutional layer is used to determine the weights.  

\section{Datasets}
\label{sec:dataset}

In the literature, three different types of events, including social, sports, and natural disaster events, are explored. In this work, we evaluate the proposed methodology on three different datasets, namely (i) social events dataset, (ii) sports event dataset, (iii) and natural disasters dataset. The basic motivation behind the evaluation of the proposed work on these datasets is to analyze how DL models respond to different types of events covering diversified visual content. In the next subsections, we provide a detailed description of the datasets.

\subsection{Social Events Dataset}
\begin{figure}[h]
%\label{fig:taxonomy}
\centering
\includegraphics[width=0.75\textwidth]{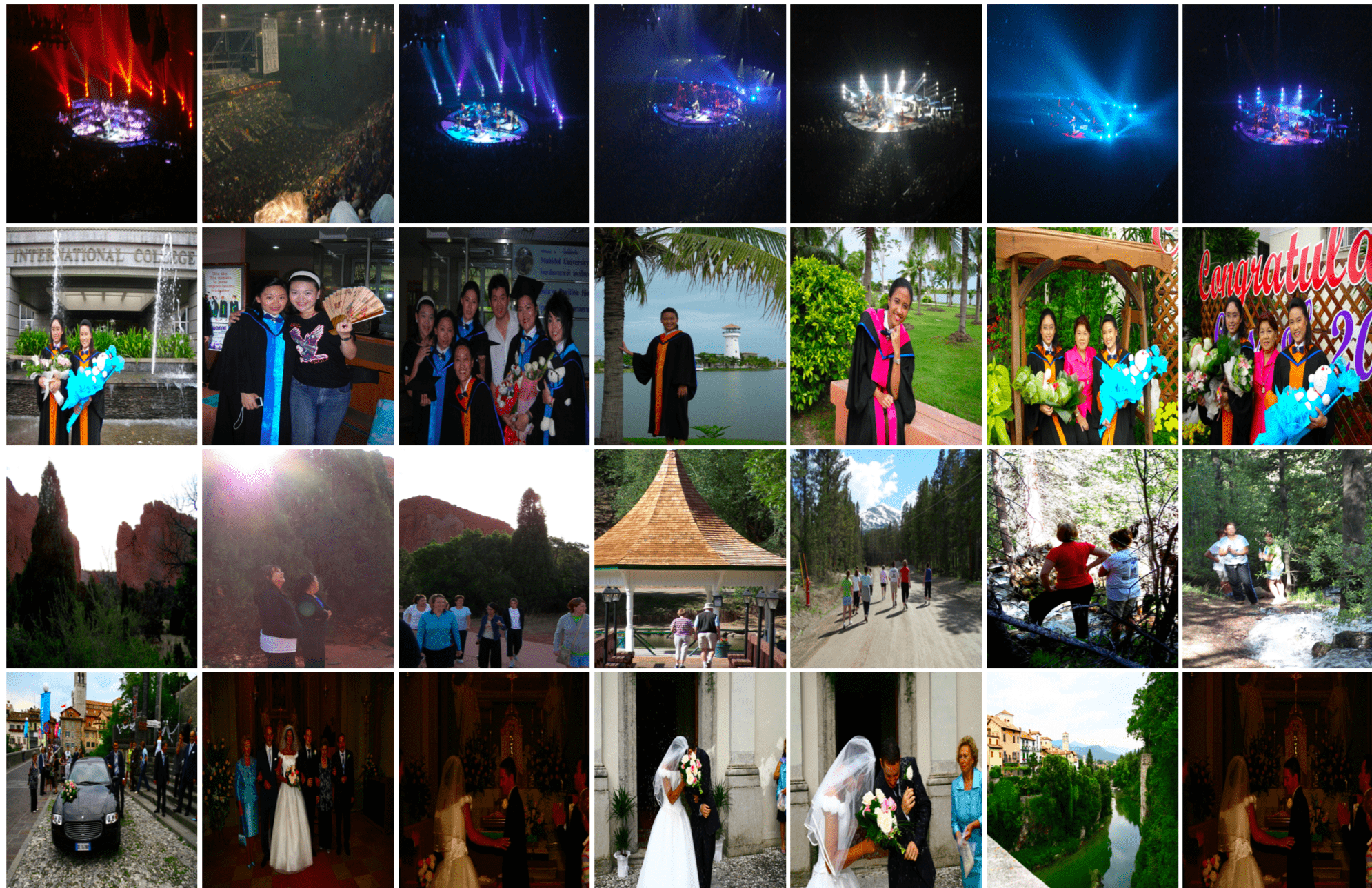}
\caption{Sample images from the social events dataset.}
	\label{fig:sample_1}
\end{figure}

Social events are among the most commonly used events in the literature \cite{ahmad2019deep}. Several datasets, such as USED \cite{ahmad2016used} and EiMM \cite{mattivi2011exploitation} covering different types of social events, such as wedding, graduation, holiday trips, etc., are publicly available for research purposes. In this work, experiments are conducted on the USED dataset. The dataset is provided into two subsets one containing $6$ different types of social events while the other cover images from $8$ types of events. We consider the 6 events including \textit{concert}, \textit{graduation}, \textit{mountain trip}, \textit{picnic}, \textit{sea holiday}, and \textit{ski holiday}. Some sample images from the dataset are provided in Figure \ref{fig:sample_1}.

%%%%%%%%%%%%%%%%%%%%%%%%
%%%%%%%%%%%%%%%%%%%%%%%%%%

%%%%%%%%%%%%%% FIGURE %%%%%%%%%%%%%%%%%%%%%%%%%%%%%%%%%%%%%%%

\begin{figure}[h]
%\label{fig:taxonomy}
\centering
\includegraphics[width=0.75\textwidth]{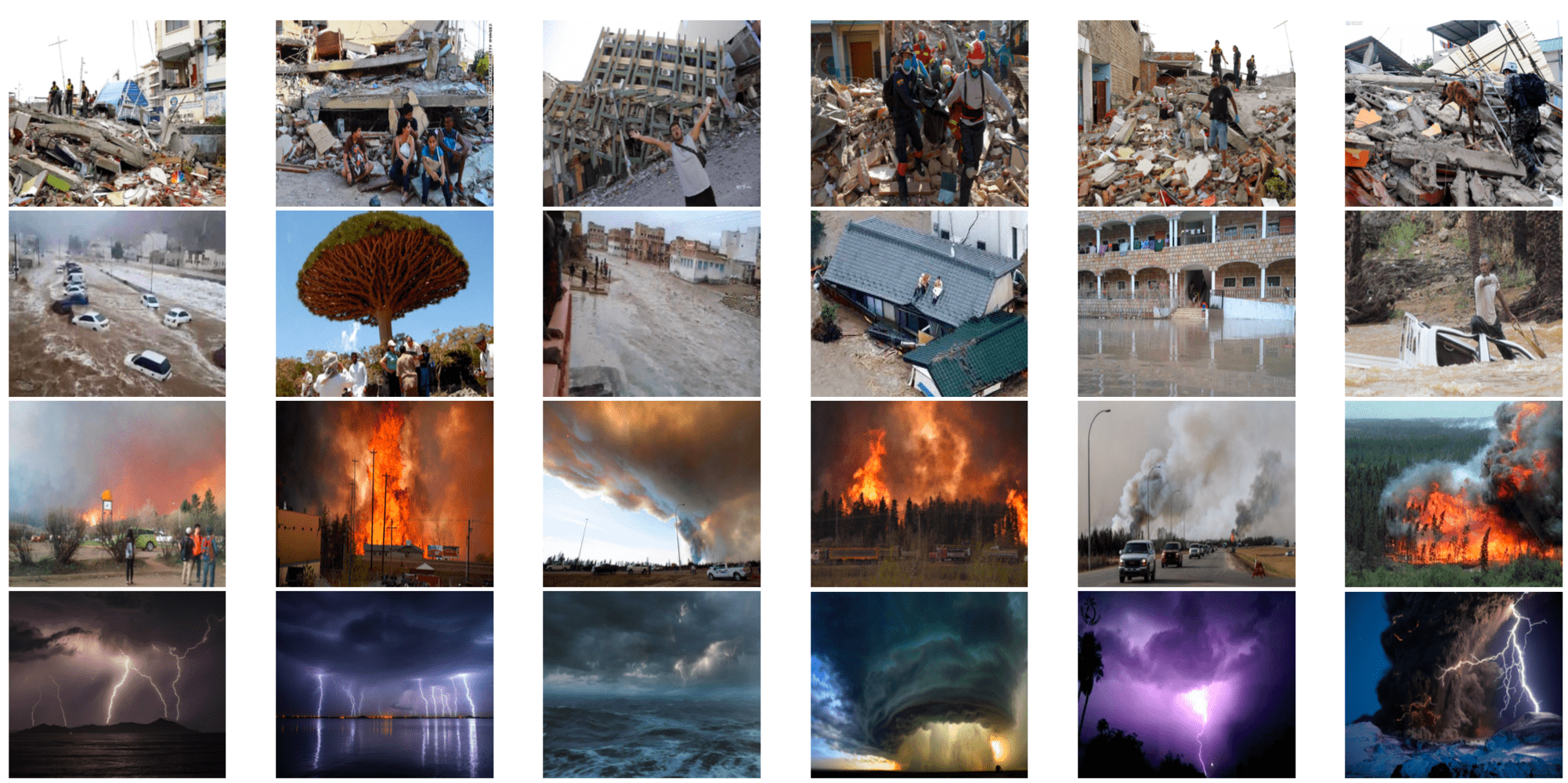}
\caption{Some sample images from natural disaster dataset.}
	\label{fig:sample_2}
\end{figure}
%%%%%%%%%%%%%%%%%%%%%%%%%%%%%%%%%%%%%%%%%%%%%%%%%%%%%%%%%%%%%
\begin{figure}[h]
%\label{fig:taxonomy}
\centering
\includegraphics[width=0.75\textwidth]{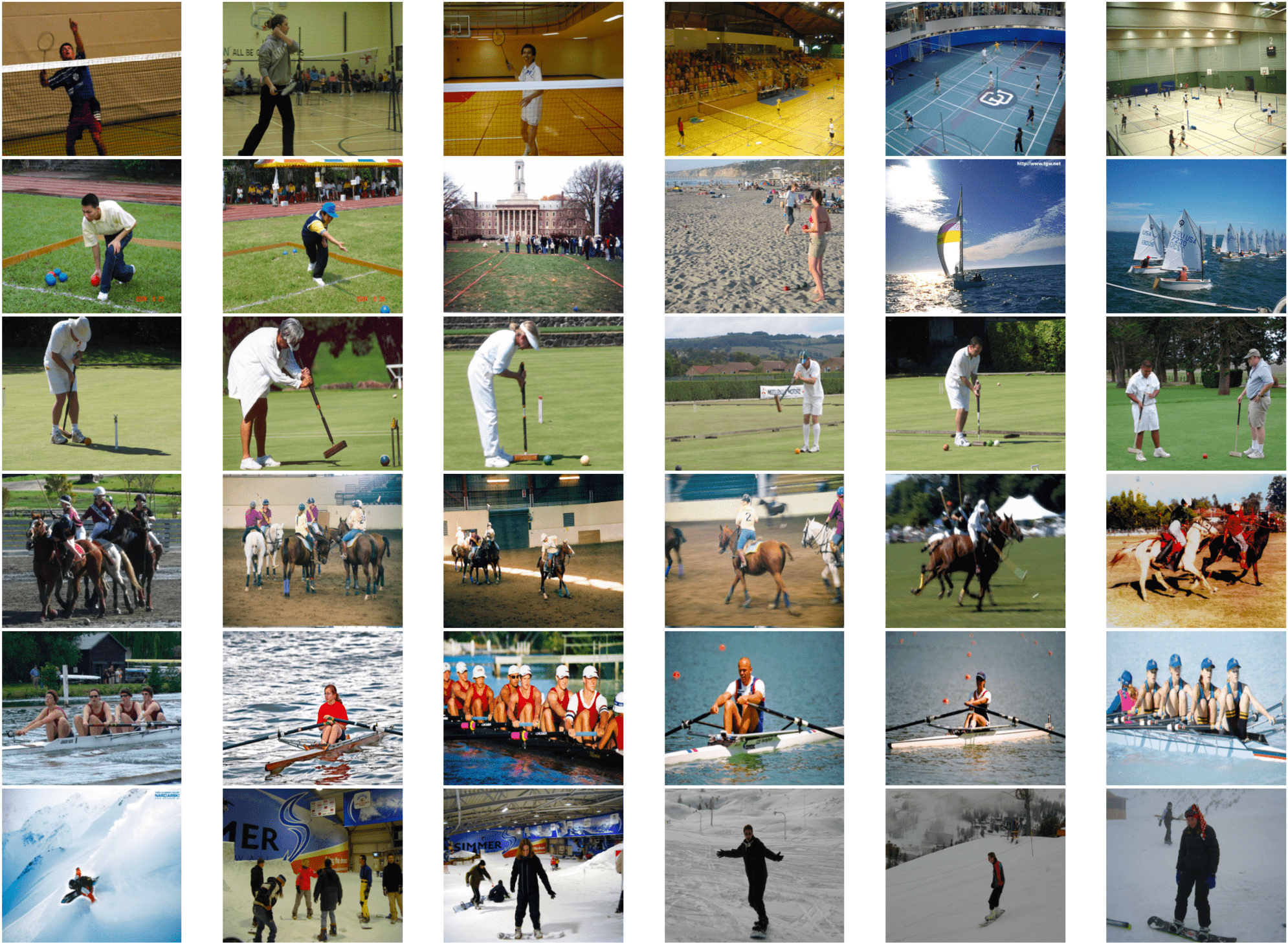}
\caption{Some sample images from sports events.}
	\label{fig:sample_3}
\end{figure}

%%%%%%%%%%%%%%%%%%%%%%%%%%%%%%%%%%%%%%%%%%%%%%

\subsection{Sports Events Dataset}
Sports events are also widely explored in the literature. To this aim, mostly UIUC Sports Events Dataset \cite{li2007and} and EiMM datasets \cite{mattivi2011exploitation} have been used. The UIUC sports events dataset is composed of $8$ events including \textit{rowing}, \textit{badminton}, \textit{polo}, \textit{bocce}, \textit{snowboarding}, \textit{croquet}, \textit{sailing}, and \textit{rock climbing}. Besides the class labels, the images are also categorized in terms of difficulty level and the distance of object and for-grounds. Two different difficulty levels, namely (i) easy, (ii) medium are defined on the basis of annotators' judgment. Both the training and test provide equal contribution from each category of images. 

On the other hand, the EiMM sports dataset contains images from $9$ different types of sports events, namely \textit{baseball}, \textit{basketball}, \textit{bike}, \textit{cycling}, \textit{formula 1}, \textit{golf}, \textit{hockey}, \textit{rowing}, \textit{skating}, and \textit{swimming}. All the images are downloaded from Picasa Web Album service. The annotation is provided at two different levels including events and sub-events labels covering two different tasks, namely (i) event recognition/classification, and (ii) sub-event detection in an image. In this work, we considered the event classification task only. We note that in this work EiMM data is used for the experiments. Figure \ref{fig:sample_3} provides some sample images from the dataset.

\subsection{Natural Disaster Events Dataset}
Natural disaster analysis is relatively a new domain and still lacks a large-scale benchmark dataset. In this work, we utilize a subset of the self-collected dataset for our earlier work on natural disaster analysis \cite{ahmad2019social,ahmad2018comparative}. The dataset is composed of $4$ common natural disasters, namely \textit{earthquakes}, \textit{floods}, \textit{thunderstorms}, and \textit{wildfires}. Compared to social and sports events, natural disaster-related images provide more diversified contents, and thus more challenging. Figure \ref{fig:sample_2} provides some sample images from the dataset. 

\section{Experiments and Results}
\label{sec:experments}
In this section, we provide details of the experimental setup, conducted experiments, and results. 

\subsection{Experimental Setup}
We made several changes for fine-tuning an existing pre-trained model on our datasets. As a first step, a normalization layer is used to adjust the input in the range of $(-1., +1.)$ as required for using the pre-trained model's weights. Moreover, a very low learning rate (i.e., 1e-5) is used for the lower layers of the models to learn slowly compared to the newly added layers. The newly added layers include a global averaging pooling layer and a regular densely connected layer. Moreover, activation maps are generated from the last convolutional layer namely ``block14\_sepconv2\_act``. Moreover, Adam optimizer is used for 10 epochs with a batch size of 120. 

\subsection{Experimental Results}

\subsubsection{Natural Disasters Events}
Table \ref{tab:disasters_results} provides the experimental results of the model on the natural disaster dataset in terms of precision, recall, and F1-score. As can be seen in the table, overall good results are obtained with a weighted average F1-score of 0.91\%. 

%%%%%%%% Natural Disaster Results %%%%%%%%%
\begin{table}[]
\centering
 \caption{Experimental results on natural disasters events in terms of precision, recall, and F1-score.}
 \label{tab:disasters_results}
\begin{tabular}{|c|c|c|c|}
\hline
\textbf{Class} & \textbf{Precision} & \textbf{Recall} & \textbf{F1 Score} \\ \hline
Earthquake & 0.91 & 0.92 & 0.91 \\ \hline
 Floods&  0.84 & 0.94 & 0.89 \\ \hline
Thunder Storm & 0.89 &0.83  & 0.86 \\ \hline
Wildfires & 0.98 & 0.94 & 0.96 \\ \hline
Weighted Average & 0.91 & 0.91 & 0.91 \\ \hline
\end{tabular}
\end{table}

%%%%%%%%%%%%%%%%%%%%%%%%%%%%%%%%%%%%%%%%%

\begin{figure}[h]
%\label{fig:taxonomy}
\centering
\includegraphics[width=0.75\textwidth]{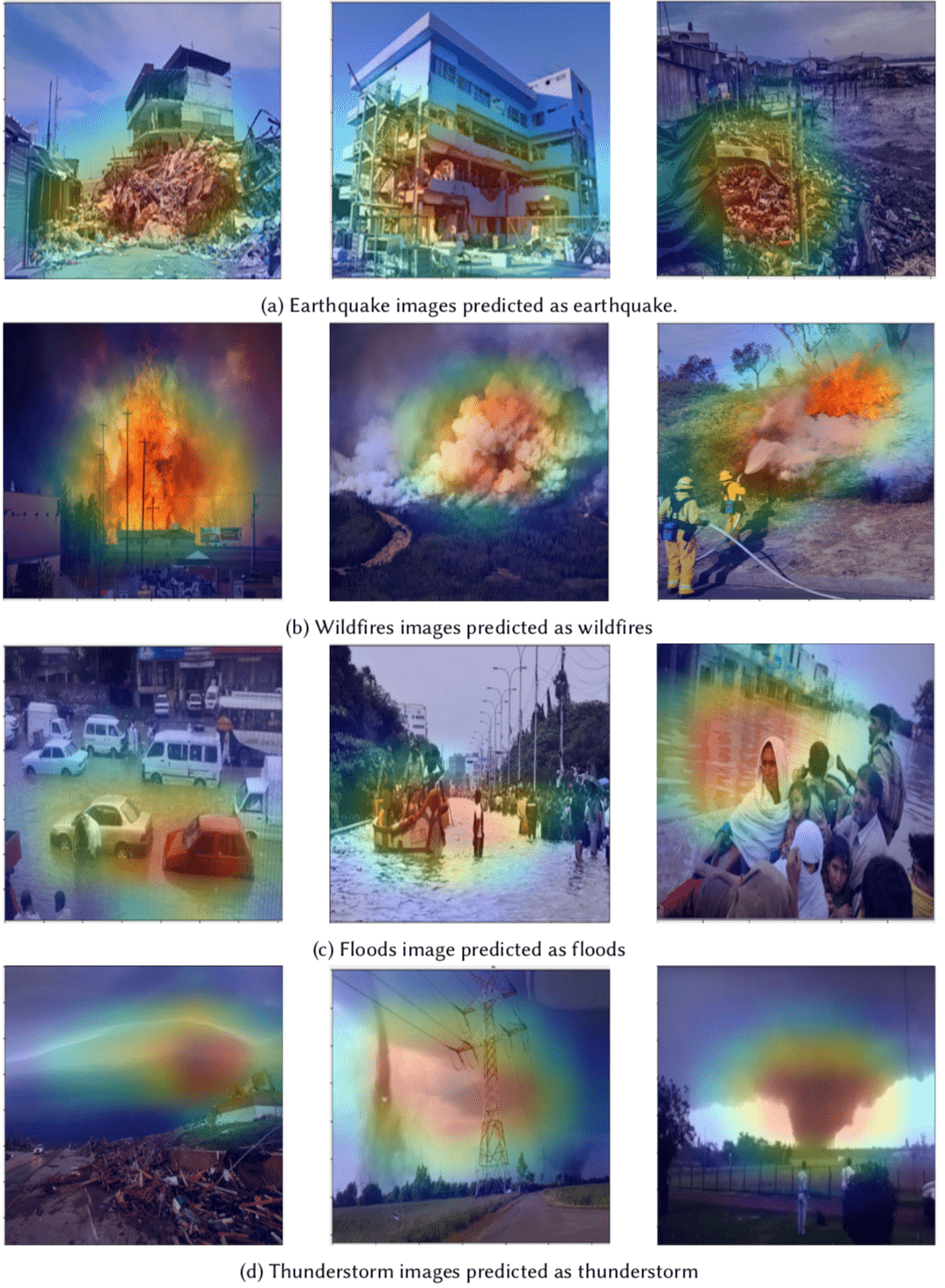}
\caption{Activation maps generated via Grad-CAM of correctly classified samples from natural disaster events dataset.\textit{As can be seen, in most of the cases, the model decisions are made on correct objects/regions.}}
	\label{fig:correct_samples_disaster}
\end{figure}

The high scores on each type of natural disaster show the generalization capabilities of the model. However, it would be interesting to see if the model's decision is based on event-related objects or not? and are the objects identified by the model as distinguishing features for the classes make sense to humans or not? To answer these questions, as a sample, we provide the activation maps generated by the Grad-CAM algorithm in Figure \ref{fig:correct_samples_disaster}. 

As can be seen in the figure, the image regions/objects used by the model (as highlighted in the figures) are relevant to the underlying events. For instance, in the earthquake samples shown in Figure \ref{fig:correct_samples_disaster}(a) only the broken parts of the buildings are highlighted in the activation maps. Similarly, in wildfires images, only the regions containing fire or smoke are highlighted in the activation maps. The same is the case with the other samples from the dataset. 

%%%%%%%%%%%%%%%%%%%%%%%%%%%%%%%%%%%%%%%%%%%%%%%%%%%%%%%%%%%%%
%%%%%%%%%%%%%%%%%%%%%%%%%%%%%%%%%%%%%%%%%%%%%%%%%%%%%%%%%%%%%

%%%%%%%%%%%%%%%%%%%%%%%%%%%%%%%%%%%%%%%%%%%%%%

\begin{figure}[h]
%\label{fig:taxonomy}
\centering
\includegraphics[width=0.75\textwidth]{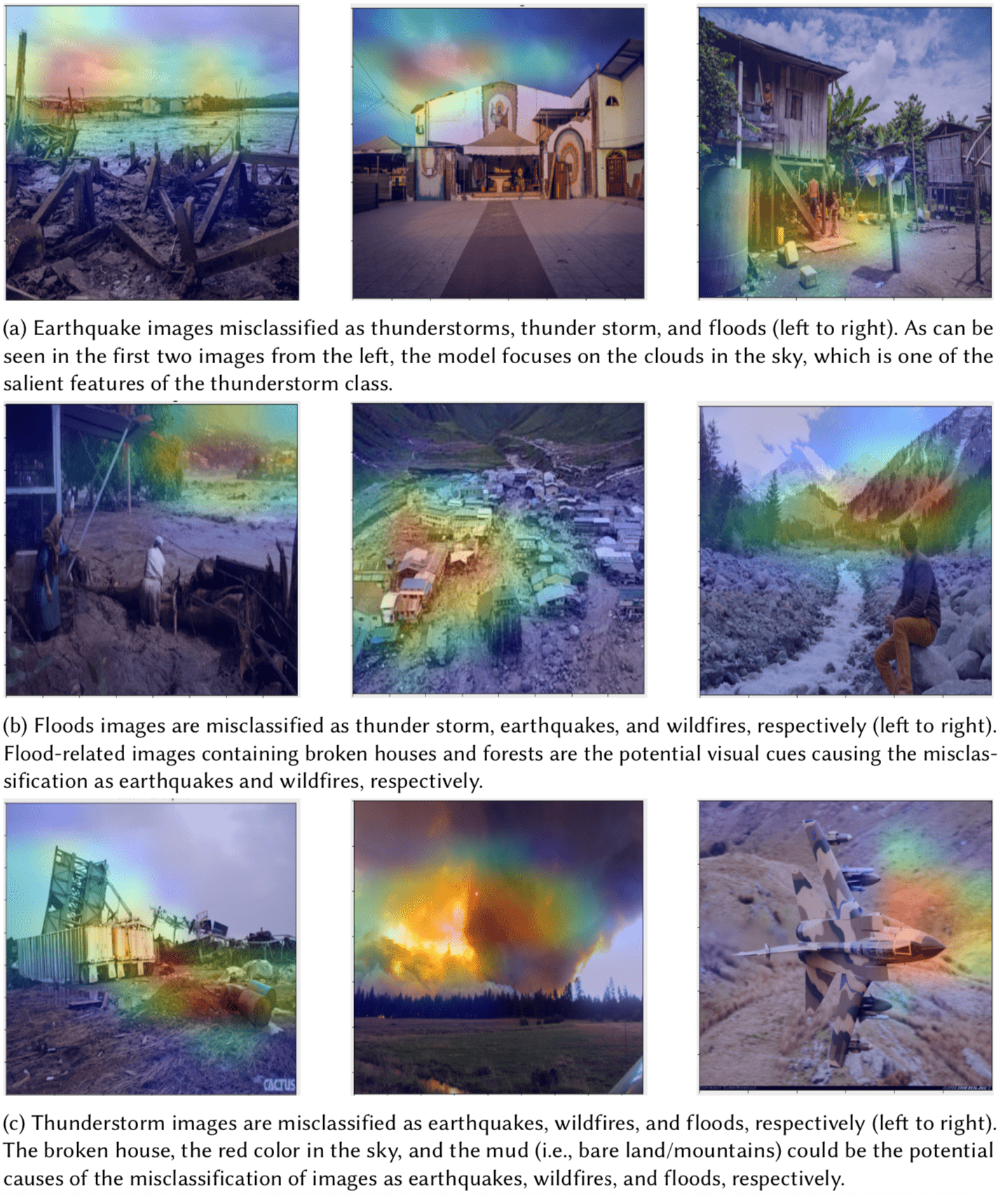}
\caption{Wrongly predicted samples from natural disasters events Dataset. \textit{The misclassification is mostly due to visual similarities with the classes.}}
	\label{fig:wrong_samples_disaster}
\end{figure}

It is also important to analyze the activation maps of the misclassified samples. This will help us in understanding the causes of the misclassification. In Figure \ref{fig:wrong_samples_disaster}, we provide some misclassified samples from different classes. The activation maps of the misclassified samples help in analyzing the potential causes of the model failure. Apparently, the misclassification seems due to the similar type of texture/content of the images with the images from the misclassified classes. For instance, as can be seen in Figure \ref{fig:wrong_samples_disaster}(a), in the first two images (from the left) the model focuses on the clouds in the sky,  which is one of the salient features of the thunderstorm class, instead of the broken house. One of the potential reasons could be the less or un-noticeable damage to the house as the building seems undamaged in the second image. Thus, despite the failure in correctly predicting the events, the decision of the model makes sense to human observers.  
%%%%%%%%%%%%%%%%%%%%%%%%%%%%%%%%%%%%%%%%%%%%
%%%%%%%%%%%%%%%%%%%%%%%%%%%%%%%%%%%%%%%%%%%%%%%%%%%%%%%%%%%%%

%%%%%%%%%%%%%%%%%%%%%%%%%%%%%%%%%%%%%%%%%%%%%%
\subsubsection{Sports Events}
Table \ref{tab:sports_results} provides the experimental results of the model on the sports events dataset. As can be seen, the overall weighted average as well as precision, recall, and F1-score on the individual classes is very high. 

%%%%%%%% Sports Events Results %%%%%%%%%
\begin{table}[]
\centering
 \caption{Experimental results on sports events in terms of precision, recall, and F1-score.}
 \label{tab:sports_results}
\begin{tabular}{|c|c|c|c|}
\hline
\textbf{Class} & \textbf{Precision} & \textbf{Recall} & \textbf{F1 Score} \\ \hline
 Baseball& 0.94 & 0.98 & 0.96 \\ \hline
 Basketball & 0.96 & 0.95 & 0.95 \\ \hline
  Bike & 0.94 & 0.91 & 0.93 \\ \hline
  Cycling  & 1.0 & 0.89 & 0.94 \\ \hline
 F1 Race   & 0.88 & 0.98 & 0.93 \\ \hline
Golf & 0.98 & 0.99 & 0.98 \\ \hline
 Hockey   & 0.96 & 0.99 & 0.98 \\ \hline
Rowing & 0.99 & 0.99 & 0.99 \\ \hline
  Skating  & 0.97 & 0.99 & 0.98 \\ \hline
Swimming & 0.99 & 0.99 & 0.99 \\ \hline
Weighted Average & 0.97 & 0.97 & 0.97 \\ \hline
\end{tabular}
\end{table}

%%%%%%%%%%%%%%%%%%%%%%%%%%%%%%%%%%%%%%%%%

\begin{figure}[h]
%\label{fig:taxonomy}
\centering
\includegraphics[width=0.75\textwidth]{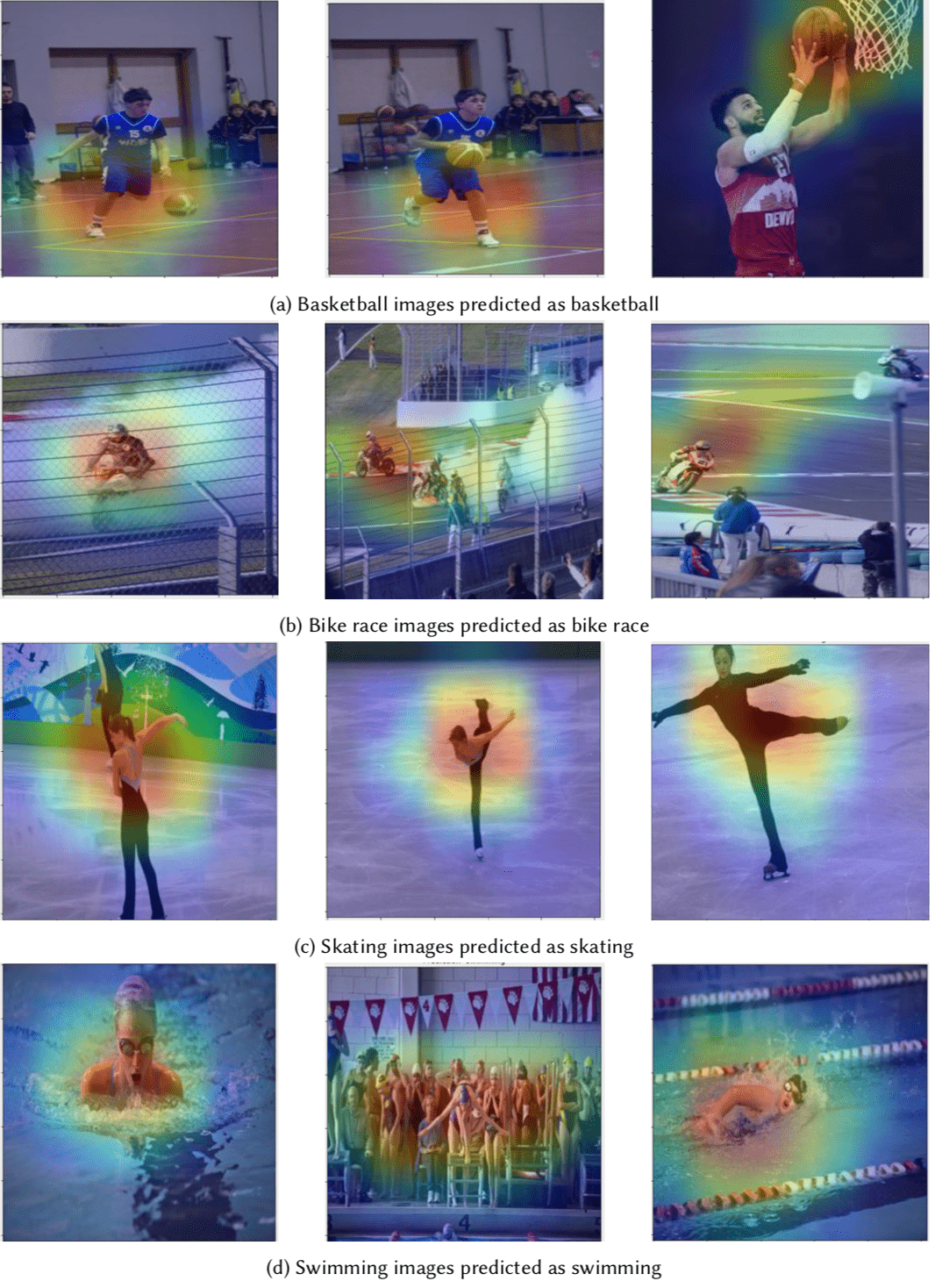}
\caption{Activation maps generated via Grad-CAM of correctly classified samples from sports events dataset.\textit{As can be seen, the model decisions are made on correct objects/regions of the images.}}
	\label{fig:correct_samples_sports}
\end{figure}

Similar to natural disaster events, we also provide some sample activation maps generated through Grad-CAM in Figure \ref{fig:correct_samples_sports} to show the key objects and image regions influencing the model's decision. Similar to natural disaster events, in most cases the model's decisions are based on relevant objects. For instance, as can be seen in Figure \ref{fig:correct_samples_sports}(a), basketball-related images are classified on the basketball goods, such as the ball and basket. Similarly, the bike race events are classified on the presence of bike and rider. The same trend is also observed in the other events.
%%%%%%%%%%%%%%%%%%%%%%%%%%%%%%%%%%%%%%%%%%%%%%%%
%%%%%%%%%%%%%%%%%%%%%%%%%%%%%%%%%%%%%%%%%%%%%%%%%%%%%%%%%%%%%

\begin{figure}[h]
%\label{fig:taxonomy}
\centering
\includegraphics[width=0.75\textwidth]{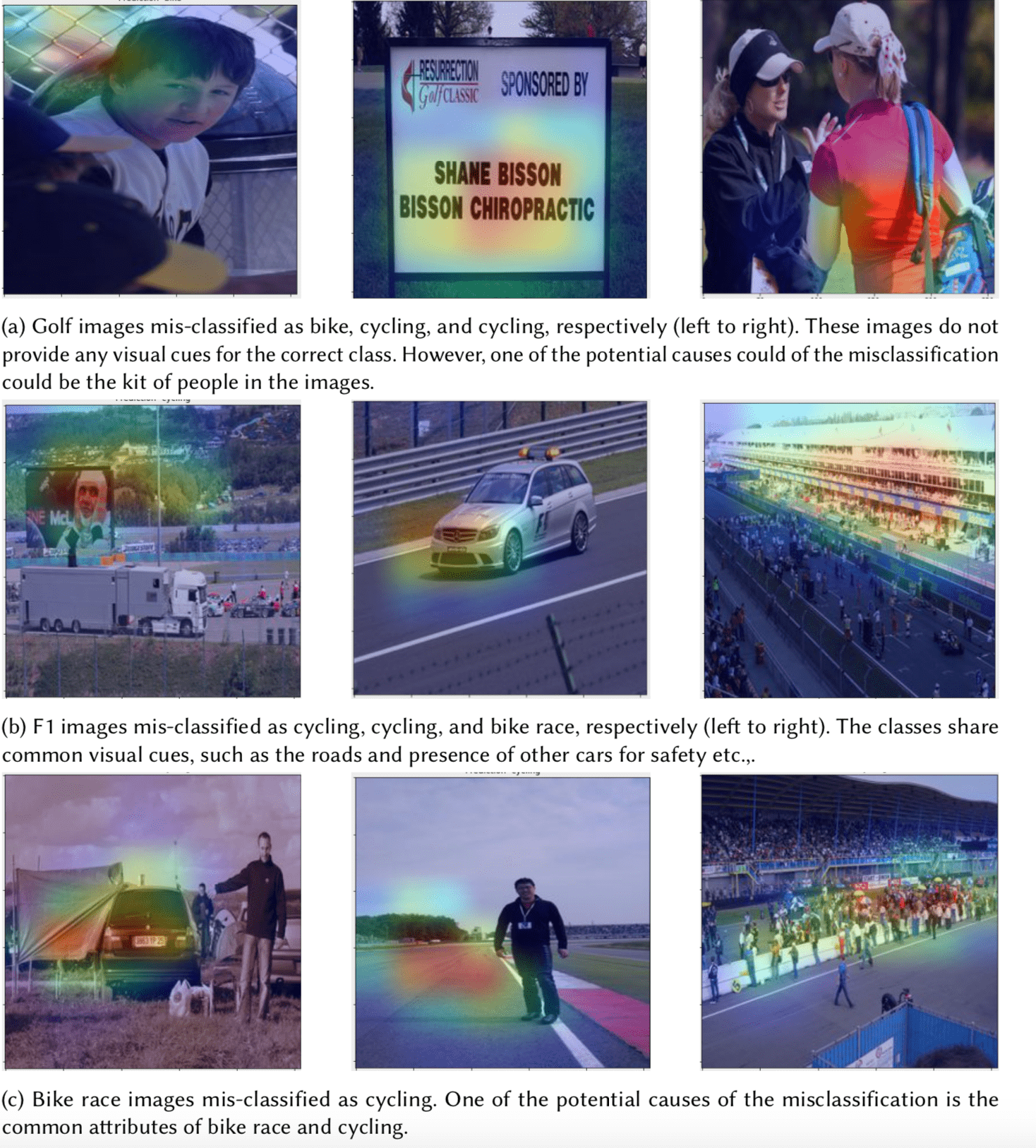}
\caption{Wrongly predicted samples from sports events Dataset. \textit{The misclassification is mostly due to visual similarities with the classes.}}
	\label{fig:wrong_samples_sports}
\end{figure}

%%%%%%%%%%%%%%%%%%%%%%%%%%%%%%%%%%%%%%%%%%%%%%
The activation maps of the misclassified samples from the sports events dataset are provided in Figure \ref{fig:wrong_samples_sports}. As can also be seen in the figure, most of the misclassified samples lack event salient objects, such as bike for bike racing, cycle for cycling, etc. Thus the decisions seem to have been taken on the basis of secondary i.e, less relevant objects, which is one of the potential causes of the misclassification.

%%%%%%%%%%%%%%%%%%%%%%%%%
%%%%%%%%%%%%%%%%%%%%%%%%%%%%%%%%%%%%%%%%%%%%%%%%%%%%%%%%%%%%%

%%%%%%%%%%%%%%%%%%%%%%%%%%%%%%%%%%%%%%%%%%%%%%

\subsubsection{Social Events}
Table \ref{tab:social_results} provides the experimental results on the social events dataset. Similar to the other datasets, the model has shown outstanding recognition capabilities on the dataset by obtaining an overall weighted average F1-score of 0.94. The high score on the dataset shows the better performance of the model in differentiating among the social events. However, it will be interesting to see whether the predictions made on the relevant object or not?  

%%%%%%%% Social Events Results %%%%%%%%%
\begin{table}[]
\centering
 \caption{Experimental results on social events in terms of precision, recall, and F1-score.}
 \label{tab:social_results}
\begin{tabular}{|c|c|c|c|}
\hline
\textbf{Class} & \textbf{Precision} & \textbf{Recall} & \textbf{F1 Score} \\ \hline
 Concert& 0.93 & 0.94 & 0.93 \\ \hline
 Graduation & 0.96 & 0.92 & 0.94 \\ \hline
  Mountaintrip & 0.91 & 0.95 & 0.93 \\ \hline
Picnic  & 0.94 & 0.92 & 0.93 \\ \hline
 Sea Holidy   & 0.95 & 0.91 & 0.93 \\ \hline
Ski Holiday & 0.93 & 0.98 & 0.95 \\ \hline
Weighted Average & 0.94 & 0.94 & 0.94 \\ \hline
\end{tabular}
\end{table}

%%%%%%%%%%%%%%%%%%%%%%%%%%%%%%%%%%%%%%%%%

\begin{figure}[h]
%\label{fig:taxonomy}
\centering
\includegraphics[width=0.75\textwidth]{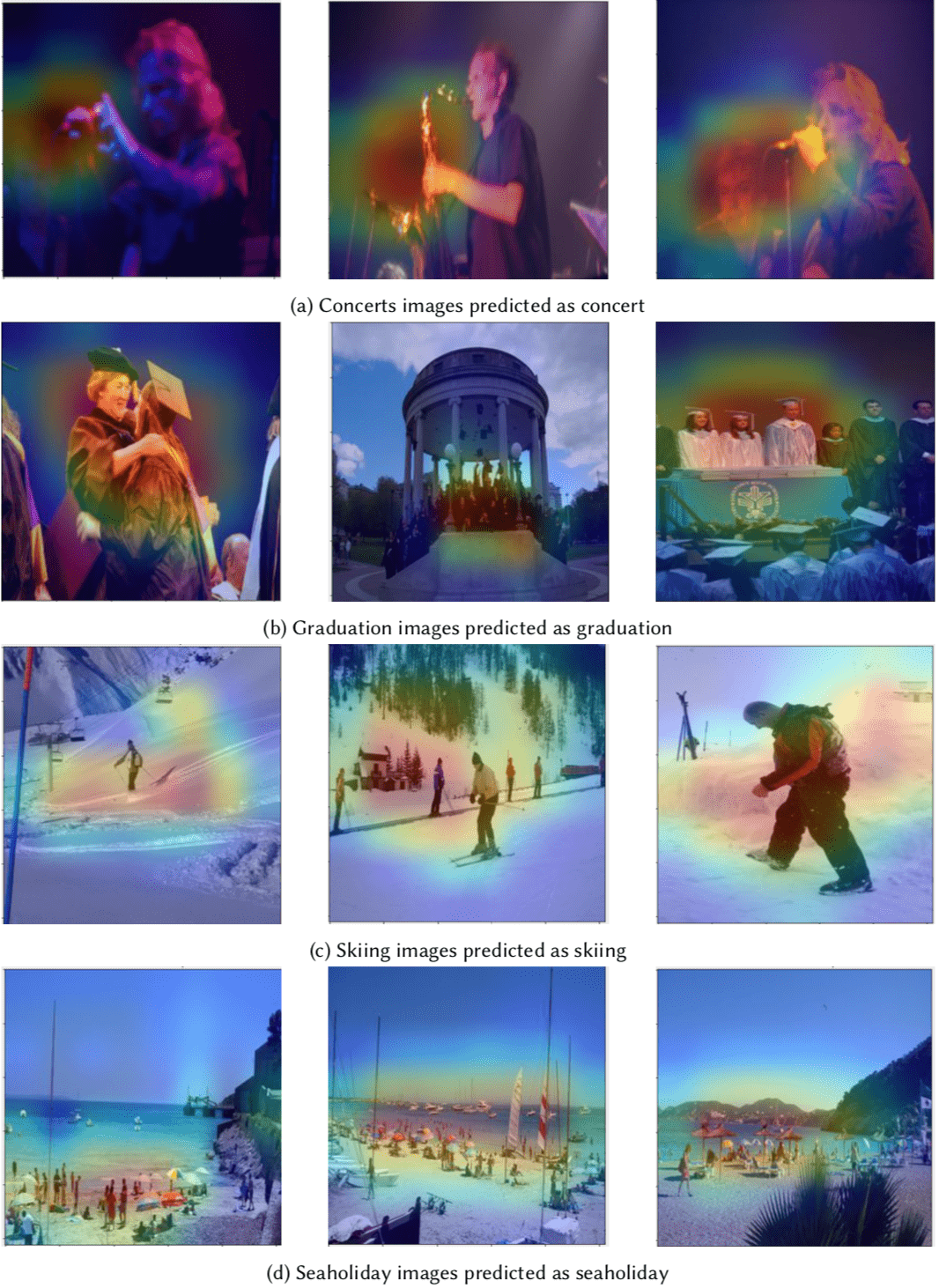}
\caption{Activation maps generated via Grad-CAM of correctly classified samples from social events dataset.\textit{As can be seen, the model decisions are made on correct objects/regions of the images.}}
	\label{fig:correct_samples_social}
\end{figure}

Figure \ref{fig:correct_samples_social} provides some sample activation maps generated on social events images. Similar to natural disasters and sports events, in most cases, the predictions are made on the basis of event-related objects. For instance, as can also be seen in Figure \ref{fig:correct_samples_social}(a), concerts images are classified on the presence of musical instruments and lighting. Similarly, the graduation images are classified on the basis of the graduation gown and cap.

%%%%%%%%%%%%%%%%%%%%%%%%%%%%%%%%%%%%%%%%%%%%%%%%%%%%%%%%%%%%%

\begin{figure}[h]
%\label{fig:taxonomy}
\centering
\includegraphics[width=0.75\textwidth]{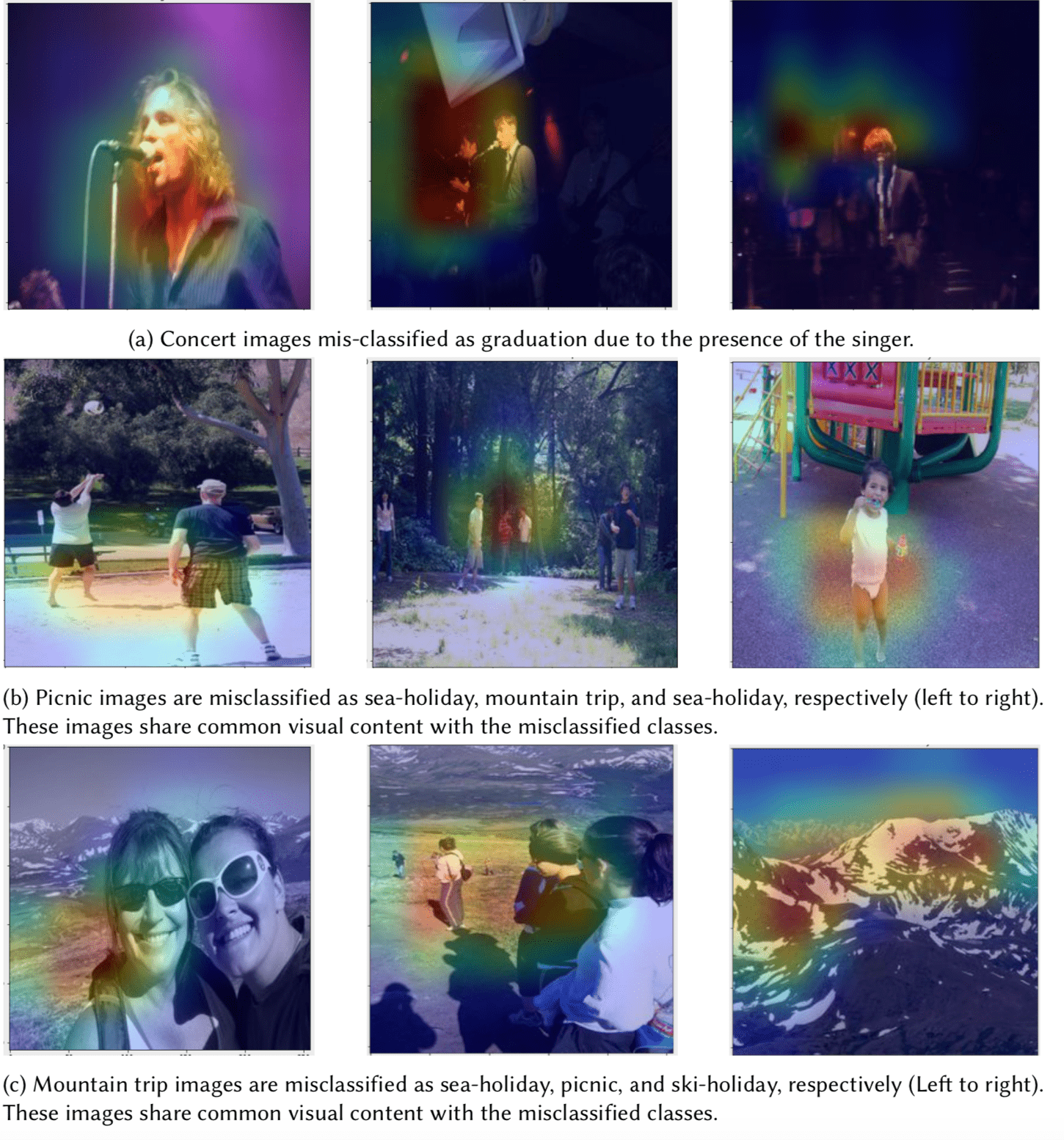}
\caption{Wrongly classified samples from social events dataset.\textit{The misclassification is mostly due to visual similarities with the confused classes.}}
	\label{fig:wrong_samples_social}
\end{figure}

%%%%%%%%%%%%%%%%%%%%%%%%%%%%%%%%%%%%%%%%%%%%%%
Figure \ref{fig:wrong_samples_social} provides some misclassified samples from the social events dataset to analyze the potential causes of the model's failure. As can also be seen in Figure \ref{fig:wrong_samples_social}(a), the concert images are mostly misclassified as graduation mainly due to the presence of the singer. We found several misclassified images where a singer in front of a mic has been highlighted as the object influencing the model's decision. Similarly, picnic images are mostly confused with sea-holiday, mountain trip, and ski-holiday due to the common visual cues. The same is the case with the other classes. 

%%%%%%%%%%%%%%%%%%%%%%%%%%%%%%%%%%%%
%%%%%%%%%%%%%%%%%%%%%%%%%
%%%%%%%%%%%%%%%%%%%%%%%%%%%%%%%%%%%%%%%%%%%%%%%%%%%%%%%%%%%%%

%%%%%%%%%%%%%%%%%%%%%%%%%%%%%%%%%%%%%%%%%%%%%%
\subsection{Subjective Analysis through Crowd-sourcing} 
In the subjective analysis, we aim to manually analyze the activation maps in a crowd-sourcing study to verify whether the decision of the ML model is based on event-related objects/regions or not? To this aim, a group of Masters students was asked to manually check the activation maps of correctly predicted samples generated through Grad-CAM as shown in Figure \ref{fig:correct_samples_disaster}, Figure \ref{fig:correct_samples_sports}, and Figure \ref{fig:correct_samples_social}, and decide if the objects/regions on the basis of which the predictions are made belong to the underlying event or not. Each sample is manually analyzed by at least three different persons, and the final decision is made on the majority votes. The participants of the study were asked to label the samples either as $1$ or $0$, where label $1$ represents that the prediction is made on the basis of event-related objects/regions. Once all the samples are labeled, we computed a score by dividing the number of samples labeled as $1$ divided by the total number of the samples. We termed this score as the accuracy of the model in identifying and making a decision on event-related objects/regions. 

Table \ref{tab:subjective_disasters_results}, Figure \ref{tab:subjective_social_results}, and Figure \ref{tab:subjective_sports_results} show the scores obtained during the study on natural disasters, social, and sports events, respectively. Overall good scores are obtained on each dataset with a highest score of .84 on social events dataset. One of the possible reasons of the higher score on the dataset could be linked with the high inter class variation.
%%%%%%%% Natural Disaster Results %%%%%%%%%
\begin{table}[]
\centering
 \caption{Subjective analysis results on the natural disasters events in terms of accuracy.}
 \label{tab:subjective_disasters_results}
\begin{tabular}{|c|c|}
\hline
\textbf{Class} & \textbf{Accuracy} \\ \hline
Earthquake & 0.80 \\ \hline
 Floods&  0.76 \\ \hline
Thunder Storm & 0.77  \\ \hline
Wildfires & 0.81  \\ \hline
Weighted Average & 0.78 \\ \hline
\end{tabular}
\end{table}

%%%%%%%%%%%%%%%%%%%%%%%%%%%%%%%%%%%%%%%%%

%%%%%%%% Social Events Results %%%%%%%%%
\begin{table}[]
\centering
 \caption{Subjective analysis results on social events in terms of accuracy.}
 \label{tab:subjective_social_results}
\begin{tabular}{|c|c|c|c|}
\hline
\textbf{Class} & \textbf{Accuracy} & \textbf{Class} & \textbf{Accuracy} \\ \hline
 Concert& 0.86 &  Graduation & 0.88 \\ \hline
  Mountaintrip & 0.82 & Picnic & 0.77 \\ \hline
 Sea Holidy   & 0.58 & Ski Holiday & 0.77 \\ \hline
Weighted Average & 0.78 & - & - \\ \hline
\end{tabular}
\end{table}
%%%%%%%%%%%%%%%%%%%%%%%%%%%%%%

%%%%%%%% Sports Events Results %%%%%%%%%
\begin{table}[]
\centering
 \caption{Subjective analysis results on sports events in terms of accuracy.}
 \label{tab:subjective_sports_results}
\begin{tabular}{|c|c|c|c|}
\hline
\textbf{Class} & \textbf{Accuracy} & \textbf{Class} & \textbf{Accuracy} \\ \hline
 Baseball& 0.90 & Basketball & 0.93 \\ \hline

  Bike & 0.75 & Cycling & 0.78 \\ \hline
 
 F1 Race   & 0.53 & Golf & 0.64 \\ \hline

 Hockey   & 0.98 & Rowing & 0.89 \\ \hline

  Skating  & 0.96 & Swimming & 0.82 \\ \hline
Weighted Average & 0.84 & - & - \\ \hline
\end{tabular}
\end{table}

%%%%%%%%%%%%%%%%%%%%%%%%%%%%%%%%%%%%%%%%%

As far as the scores on the individual classes are concerned, the highest scores are obtained for wildfires and earthquake images while lowest scores are obtained on floods and thunder storm. Similarly, in social and sports events, the highest scores are obtained on graduation and hockey-related images, respectively. On the other hand, lowest scores in these datasets are obtained on sea holiday and F1 race. The lower scores could be linked to the complexity of these classes, however, interestingly the recall on these classes, as can be seen in Table \ref{tab:disasters_results}, \ref{tab:social_results}, and Table \ref{tab:sports_results}, are high. Since in the subjective analysis we considered the correctly predicted samples only, one of the potential reasons could be the correct prediction of complex images in these classes where the decisions are made on secondary objects due to absence of event-related objects.

%%%%%%%%%%%%%%%%%%%%%%%%%%%%%%%%%%%%%%%%%

\subsection{Lessons Learned} 

The key lesson learned from the work can be summarized as follows.

\begin{itemize}
  \item Event recognition in single images is a challenging task compared to object recognition as event-related images generally include multiple objects and scene-level details, which are not necessarily all related to the underlying event.
  \item The model has shown outstanding capabilities in predicting complex events.
    \item The concept of explanation of the ML model by generating activation maps from the final convolutional layer allows to analyze whether the model's decisions make sense to a human observer or not. 
    \item The activation maps generated via Grad-CAM in this work show that in the majority of cases the model's decisions are based on event-related objects.
    \item In most of the cases the misclassification is due  to visual similarities with the confused classes or the absence of event salient objects. 
    \item The subjective analysis indicates that generally, the model performed well in identifying event salient objects and image regions. However, the results show a significant gap, which should ideally be near to 100\% (i.e., ideally all the decisions of the models should be based on event-salient objects). The gap indicates the need for more efforts on event-salient aspects of event recognition.  
\end{itemize}

\section{Conclusion}
\label{sec:conclusion}
In this work, we presented an explainable event recognition solution to analyze whether the prediction of the model is based on event-related objects/regions or not. To this aim, the Grad-CAM algorithm is employed to generate activation maps of the images used by a CNN model. Moreover, to evaluate the performance of the model in terms of correctly identifying event-related objects/regions, a subjective study is conducted where participants were asked to analyze whether the decision is made on event-related objects or not?  Overall, the model showed better results in terms of making decisions/predictions on event-related objects. However, ideally it should be near to 100\%, and the gap indicates more efforts are required to explore this aspect of event recognition. 

\bibliographystyle{ACM-Reference-Format}
\bibliography{sample-acmsmall}
\end{document}